%% file: main.tex
\documentclass[10pt]{article}
\usepackage[T1]{fontenc}
\usepackage{arxiv}

\input{math_commands.tex}

\usepackage{booktabs}
\usepackage{graphicx}
\usepackage[font=small,labelfont={bf,color=FlashBlue},labelsep=period]{caption}
\usepackage{flafter}
\usepackage{float}
\usepackage{needspace}
\usepackage{multirow}
\usepackage{hyperref}
\hypersetup{hidelinks,pdftitle={FlashVSR+: Efficient Video Super-Resolution via Hybrid Attention and Reward-Yield Distribution Matching},pdfauthor={Ruchang Yao et al.}}
\usepackage{url}
\usepackage{wrapfig}

\usepackage{listings}

\title{FlashVSR+: Efficient Video Super-Resolution via Hybrid Attention and Reward-Yield Distribution Matching}

\author{%
  \textbf{Ruchang Yao$^{1,2}$ \quad Shi Guo$^{1 \dagger}$ \quad Junhao Zhuang$^{3}$ \quad Mingde Yao$^{2,4}$ \quad Xin Cai$^{1,2}$ }\\[0.25em]
  \textbf{Jiahao Zhan$^{2}$ \quad Bin Fu$^{1}$  \quad Xinyuan Chen$^{1}$ \quad
  Yu Qiao$^{1}$ \quad Tianfan Xue$^{1,2,4 \dagger}$}\\[0.45em]
  \normalfont $^{1}$Shanghai AI Lab \quad $^{2}$MMLAB CUHK \quad $^{3}$Tsinghua University \quad $^{4}$CPII under InnoHK\\[0.25em]
  \normalfont\footnotesize $^{\dagger}$ denotes corresponding authors.%
}

\newcommand{\flashtableformat}{%
  \small
  \setlength{\tabcolsep}{2pt}%
  \renewcommand{\arraystretch}{1.08}%
}

\begin{document}
\raggedbottom

\maketitle

\begin{abstract}
Recent video super-resolution (VSR) models achieve efficient one-step generation through diffusion distillation. However, high-resolution inference remains costly, and distilled models are difficult to align with human preferences. These challenges call for improvements in both model architecture and training strategy. We present \textbf{FlashVSR+} to address both limitations. On the architecture side, we introduce hybrid attention based on the strong locality observed in pretrained VSR models, while retaining sparse global interaction for long-range consistency. It achieves up to $2\times$ DiT speedup and 45\% lower peak memory, enabling spatially untiled 4K VSR on a single 24GB RTX 4090. On the training side, we propose Reward-Yield Distribution Matching Distillation (Ry-DMD) for preference-aligned one-step generation. Ry-DMD uses reward-weighted teacher rollouts to incorporate black-box rewards into distribution matching, without backpropagating through the reward model. We further show that Ry-DMD generalizes beyond VSR to T2I generation. Together, FlashVSR+ delivers stronger restoration quality and substantially improves high-resolution efficiency, with consistent gains in user preference. We release our code, models, and training data at \href{https://www.flashvsr.tech}{\textcolor{FlashBlue}{\textbf{www.flashvsr.tech}}}.


\end{abstract}

\begin{figure}[!ht]
\centering

\includegraphics[width=0.98\linewidth]{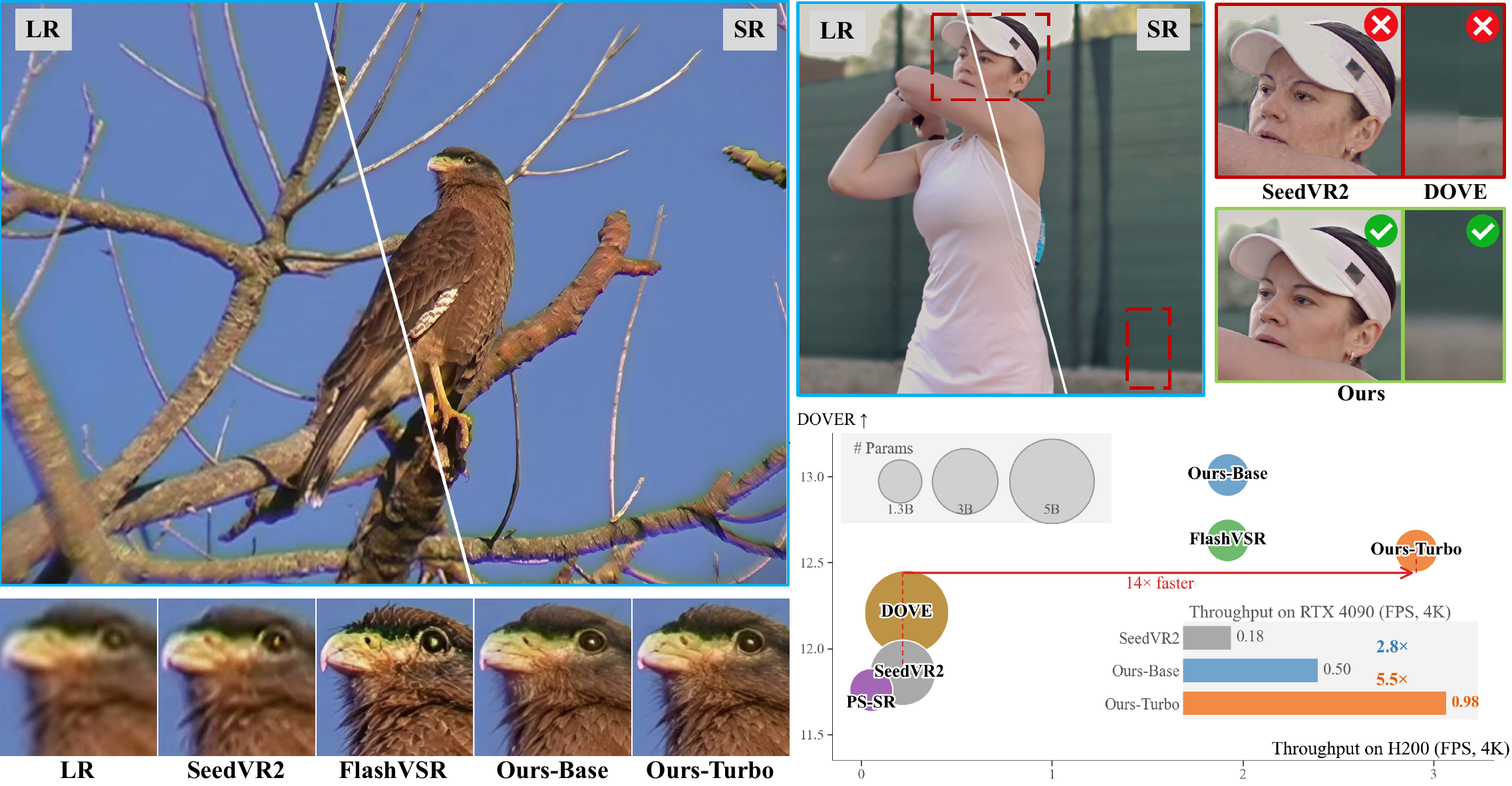}
\vspace{-0.5em}
\caption{
\textbf{FlashVSR+ achieves high-quality and efficient video restoration.}
\textbf{Left:} We recover fine visual details while maintaining natural textures.
\textbf{Right:} At 4K resolution, we suppress visible artifacts and enable substantially faster inference on both H200 equivalent GPU and a RTX 4090.
}
\label{fig:teaser}
\end{figure}

\section{Introduction}
Video super-resolution (VSR) reconstructs high-resolution videos from low-resolution inputs. It supports applications such as smartphone photography, social media, live streaming, and cost-effective upscaling of AI-generated videos to 2K or 4K resolution. Reconstruction-based VSR aggregates complementary information across frames to improve distortion oriented fidelity~\citep{chan2022basicvsrpp,chan2022realbasicvsr,shi2022rethinking}. These methods benefit from stronger propagation, alignment, and spatiotemporal modeling, but the perception--distortion trade-off often leads to over-smoothed textures when high-frequency details are ambiguous~\citep{blau2018perception}. Generative VSR addresses this limitation by utilizing learned image or video priors to synthesize perceptually realistic details~\citep{zhou2024upscale,chen2024sateco,xu2025videogigagan,li2025diffvsr}, and has achieved substantial progress in visual quality.

Despite these advances, efficient and flexible generative VSR remains hindered by two bottlenecks. (1) \textbf{Computational efficiency}: Diffusion-based approaches rely on costly multi-step denoising trajectories~\citep{zhou2024upscale,chen2024sateco}. Although recent one-step VSR methods greatly reduce this latency by distilling the multi-step generation trajectory into a single step~\citep{song2023consistency,yin2024dmd2,yin2024dmd,sun2025dloral,zhuang2026flashvsr,lv2026duovsr}, the computation and memory costs of spatiotemporal attention remain prohibitive for high-resolution videos (\emph{e.g.}, 2K and 4K resolution). As a result, inference on consumer-grade GPUs (\emph{e.g.}, RTX 4090) often requires spatial tiling, which limits global context and may introduce boundary artifacts, as illustrated in Fig.~\ref{fig:teaser}.
(2) \textbf{Preference-aligned distillation}:  Standard one-step distillation method DMD~\citep{yin2024dmd2,yin2024dmd} matches the teacher distribution but cannot explicitly favor outputs that better align with human preferences, such as more accurate text rendering. Existing reward-guided methods typically rely on multi-step stochastic trajectories~\citep{black2024ddpo,fan2023dpok} or differentiable reward signals~\citep{clark2024draft,jia2025lasro}, making them ill-suited to deterministic one-step generators with black-box rewards. 

To address these bottlenecks, we present FlashVSR+, a one-step VSR framework with hybrid attention for efficient high-resolution restoration and Ry-DMD for preference-aligned distillation with black-box rewards. Together, they make one-step generative VSR more efficient, \textbf{enabling spatially untiled 4K VSR inference on a single 24GB RTX 4090}, while better aligning outputs with human preferences.

For high-resolution efficiency, we find that pretrained VSR models such as FlashVSR~\citep{zhuang2026flashvsr} exhibit strong spatiotemporal attention locality, but with substantial variation across layers. Attention is highly
concentrated within compact neighborhoods in many blocks, suggesting that dense attention is unnecessary throughout the backbone, while fully local attention would be overly restrictive. We therefore introduce a
\textbf{hybrid attention architecture} that replaces most original attention blocks with efficient local spatiotemporal attention while retaining a small
number of sparse global blocks. This design achieves up to $2\times$ DiT acceleration and nearly halves peak memory, enabling spatially untiled 4K VSR on a single 24GB RTX 4090 while maintaining
comparable restoration quality.

For preference-aligned distillation, one-step generators are particularly challenging because stochasticity is limited to the input seed, while black-box rewards provide no gradients for direct optimization. We propose \textbf{Reward-Yield Distribution Matching Distillation (Ry-DMD)}, which uses multiple teacher rollouts to expose the stochasticity available to a one-step generator and reweights them with black-box rewards. The difference between the reward-weighted and original teacher predictions provides a preference direction, which is added to standard DMD to favor higher-reward teacher-supported outputs. We validate Ry-DMD on OCR-guided VSR and further demonstrate its applicability to text-to-image generation.

To further scale up one-step VSR distillation, a natural strategy is to increase the training budget and data size. However, we find that longer DMD training does not yield consistent gains, revealing non-monotonic behavior. Instead, student initialization and teacher capacity play a more important role in final performance.

\section{Related Work}

\subsection{Video Super-Resolution}

Conventional VSR methods recover missing details through alignment, propagation, and spatiotemporal aggregation~\citep{wang2019edvr,chan2021basicvsr, chan2022basicvsrpp,shi2022rethinking,liang2022rvrt}, but often produce over-smoothed textures in perceptually ambiguous regions~\citep{blau2018perception}. Generative VSR instead leverages learned priors to synthesize realistic details, including adversarial~\citep{chu2020tecogan,xu2025videogigagan} and diffusion-based approaches~\citep{rota2024stablevsr,zhou2024upscale,chen2024sateco,li2025diffvsr, chen2026stcdit}. To reduce the cost of iterative sampling, recent methods increasingly target one-step restoration through direct fine-tuning, adversarial post-training, or distillation~\citep{liu2025ultravsr,chen2025dove,wang2026seedvr2, sun2025dloral,zhuang2026flashvsr,lv2026duovsr}. 
Recent image SR and restoration methods have also explored reward or preference guided post-training~\citep{yi2026gdposr,zhao2026oars,xu2025irpo}.
Nevertheless, one-step VSR still faces challenges in both high-resolution inference efficiency and reward-guided post-training. FlashVSR+ targets these two aspects with hybrid attention and Ry-DMD, respectively.

\subsection{Efficient Attention for Visual Generation}

The quadratic cost of dense attention has motivated sparse and sub-quadratic
alternatives for visual generation. Video diffusion methods exploit structured
spatiotemporal sparsity or locality-aware computation to reduce attention cost
~\citep{xi2025sparsevideogen,yang2025sparsevideogen2,
shen2025draftattention,xu2025xattention,li2025radialattention}.
Another line of work adopts efficient attention backbones, including linear
attention and hybrid architectures that periodically retain full softmax
attention for global information exchange
~\citep{xie2025sana,xie2025sana15,chen2026sanavideo,chen2026sanavideo2}.
In contrast, FlashVSR+ exploits the locality observed in a pretrained VSR model,
replacing most global-attention blocks with spatiotemporal local attention while
retaining a small number of global blocks.

\subsection{Reward Optimization for One/Few-Step Generative Models}

Reward-guided post-training has been extensively studied for multi-step diffusion models, where denoising trajectories enable policy optimization, differentiable reward backpropagation, and preference learning ~\citep{fan2023dpok,black2024ddpo,clark2024draft,wallace2024diffusiondpo,liu2025flowgrpo,zheng2026diffusionnft}.
Extending these approaches to few-step and one-step generators remains challenging. While few-step models still preserve short denoising trajectories, one-step generators directly map noise to samples and lack such intermediate optimization signals. Existing few-step approaches explore reward-guided distribution matching and related optimization strategies ~\citep{jiang2025dmdr,huang2026rtdmd, dong2026guidingdistributionmatchingdistillation}, while one-step methods typically rely on
inference-time optimization or differentiable reward surrogates
~\citep{eyring2024reno,jia2025lasro,zhu2026crd}.
Ry-DMD instead enables black-box reward optimization for deterministic one-step generators by introducing a reward-tilted teacher target within the DMD framework.

\section{Method}
\label{sec:method}

\subsection{Overview}
\label{sec:method_overview}

\begin{figure}[h]
\vspace{-0.5em}
\centering
\includegraphics[width=\linewidth]{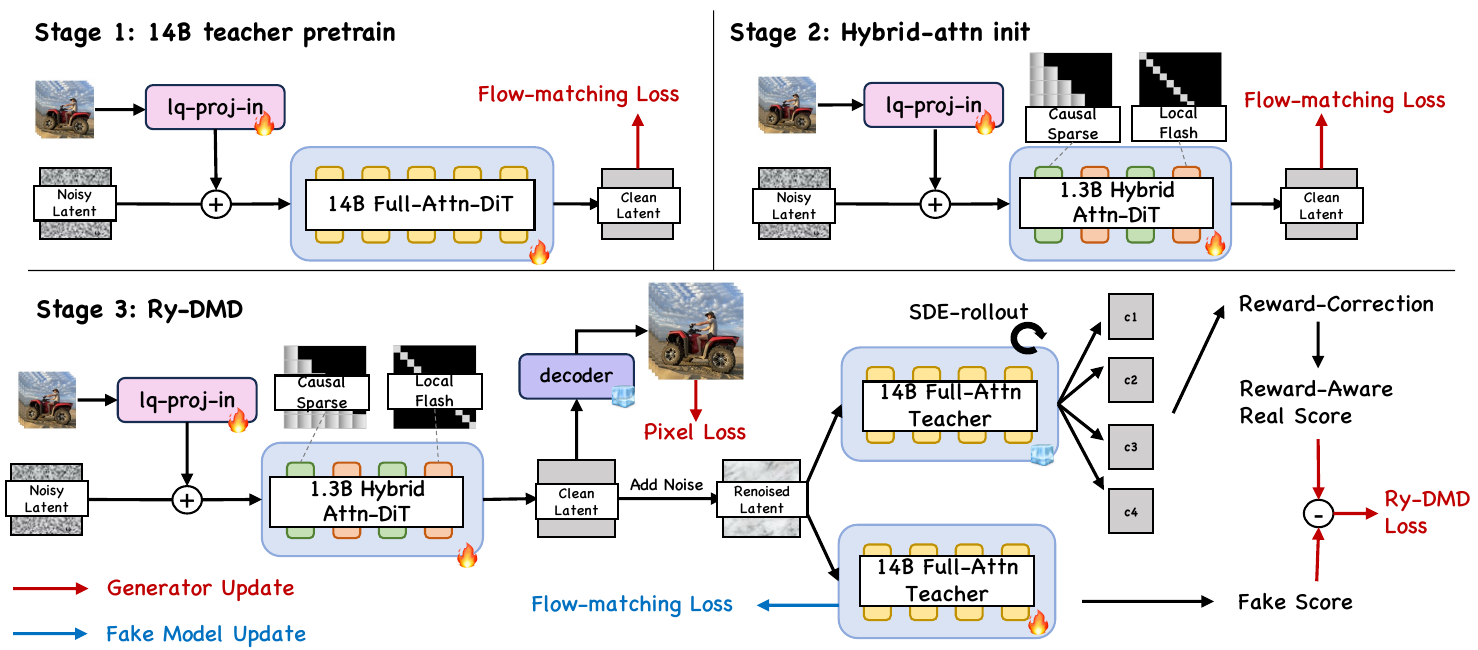}
\vspace{-1.8em}
\caption{
\textbf{Overview of the FlashVSR+ training pipeline.}
Stage 1 trains a 14B full-attention multi-step teacher on image and video data.
Stage 2 trains the hybrid-attention student to obtain a strong initialization.
Stage 3 distills the student into a one-step generator using Ry-DMD, where reward feedback modifies the teacher term estimation without requiring reward gradients.
}
\label{fig:flashvsrplus_overview}
\end{figure}

Following FlashVSR~\citep{zhuang2026flashvsr}, FlashVSR+ adopts the three-stage
training framework shown in Fig.~\ref{fig:flashvsrplus_overview}.
In \textbf{Stage 1}, we train a 14B full-attention multi-step model with flow matching as the teacher.
In \textbf{Stage 2}, we replace the 1.3B backbone with our hybrid-attention architecture and train it to provide a strong initialization for one-step generation.
In \textbf{Stage 3}, we distill the initialized student into a one-step VSR model using distribution matching together with the pixel loss. Ry-DMD further injects scalar reward feedback into the teacher term without reward backpropagation. Training details are provided in
Appendix~\ref{app:train_details}.

\Needspace{20\baselineskip}
\subsection{Efficient One-Step VSR with Hybrid Attention}
\label{sec:hybrid_attention}

\begin{wrapfigure}{r}{0.59\columnwidth}
\vspace{-1.8em}
\centering
\includegraphics[width=\linewidth]{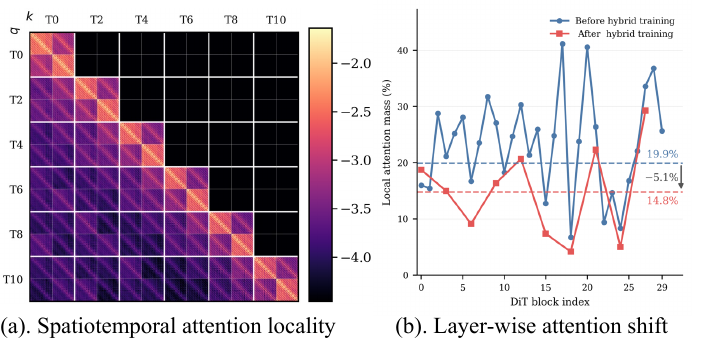}
\vspace{-1.8em}
\caption{
\textbf{Attention locality and adaptation in causal VSR.}
(a) Attention concentrates in compact spatiotemporal neighborhoods.
(b) Local attention mass varies across DiT blocks, and hybrid training reduces the locality of retained global-attention blocks.
}
\label{fig:hybrid_attention_analysis}
\vspace{-1.0em}
\end{wrapfigure}

Spatiotemporal attention is a major scalability bottleneck for causal VSR at high resolution. To understand the potential redundancy of full attention, we examine the attention patterns of a pretrained one-step VSR model. As shown in Fig.~\ref{fig:hybrid_attention_analysis}(a), attention is highly concentrated within local neighborhoods,
which cover only $\sim1\%$ of valid key blocks yet receive
$\sim20\%$ of the total attention mass. We quantify this behavior using
the local attention mass, defined as the fraction of attention assigned to valid local key blocks. Higher local attention mass indicates stronger concentration within local neighborhoods, suggesting that many non-local interactions are weakly weighted yet still require computation, leading to unnecessary overhead. As shown in Fig.~\ref{fig:hybrid_attention_analysis}(b), local attention is strongly enriched across DiT blocks but varies substantially across layers, motivating a hybrid rather than purely local design.

Motivated by this observation, we introduce \textbf{hybrid attention}, which replaces most attention blocks with efficient spatiotemporal local attention while retaining sparse global-attention blocks. Specifically, we denote the local attention configuration as $(s,t,b)$, where $s$ and $t$ denote the numbers of spatial and preceding temporal key blocks used in local attention, and $b$ denotes the number of DiT blocks replaced by local attention.
We then analyze how attention adapts after hybrid training. For visualization, we use the setting $(s,t,b)=(3,1,20)$, with after-training measurements taken only on retained global-blocks. As shown in Fig.~\ref{fig:hybrid_attention_analysis}(b), the retained global-attention blocks exhibit reduced local attention mass after hybrid training, indicating a shift toward non-local interactions. This suggests that hybrid attention allows the retained global-attention blocks to better capture non-local interactions, leading to more efficient attention computation. 

In the final model, we adopt a regular interleaving of local and global
blocks and use $(s,t,b)=(3,1,25)$, retaining one global-attention block every
five DiT blocks (Fig.~\ref{fig:flashvsrplus_overview}, Stages 2\&3).
This design achieves up to $2\times$ DiT acceleration and nearly halves peak
memory usage, with detailed evaluations in Section~\ref{sec:exp_hybrid}.

\Needspace{18\baselineskip}
\subsection{Reward-Yield Distribution Matching Distillation}
\label{sec:rydmd}

\begin{wrapfigure}{r}{0.35\columnwidth}
    \centering
    \includegraphics[width=\linewidth]{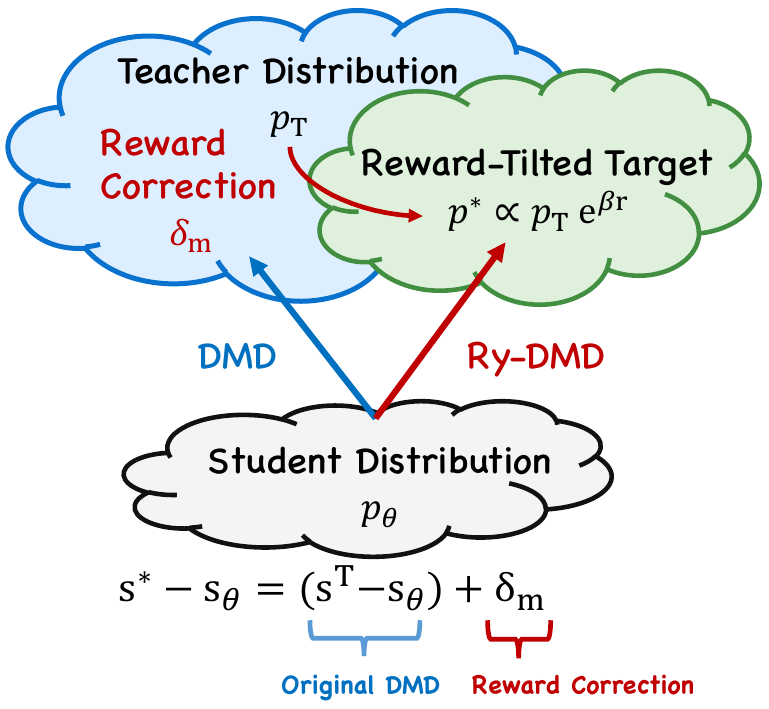}
    \caption{\textbf{Conceptual illustration of Ry-DMD.}}
    \label{fig:rydmd_overview}
\end{wrapfigure}

Adapting a deterministic one-step generator to a downstream objective or preferences is not
straightforward. Continuing standard DMD on task-specific data can adapt the
training distribution, but does not explicitly favor outputs with better human 
preference. Direct reward optimization within DMD is possible, but non-differentiable rewards require differentiable surrogates for backpropagation. We propose \textbf{Reward-Yield Distribution Matching Distillation (Ry-DMD)}. As illustrated in Fig.~\ref{fig:rydmd_overview}, Ry-DMD transforms the teacher distribution toward higher-reward outputs, such as clearer and more recognizable text, by constructing a reward-tilted target. The resulting target shift is then converted into a correction to the DMD teacher term. This preserves the overall DMD training framework while enabling the use of black-box objectives such as OCR, which we validate on VSR in Sec.~\ref{sec:exp_rydmd}; we further demonstrate its generality on text-to-image generation in Sec.~\ref{sec:exp_t2i}.

\subsubsection{Reward-Tilted Distribution Matching}

Our goal is to modify the teacher distribution toward higher-reward regions while preserving its original support.
As a distribution-level motivation, we consider an exponential tilt of the
clean teacher distribution.
Let $c$ denote the condition and $p_{\mathrm T,0}(x_0\mid c)$ the clean
teacher distribution. Given a scalar reward $r(x_0,c)$, we define the
reward-tilted distribution as
\begin{equation}
    p_0^\star(x_0\mid c)
    \propto
    p_{\mathrm T,0}(x_0\mid c)
    \exp\!\bigl(\beta r(x_0,c)\bigr),
    \label{eq:rydmd_target}
\end{equation}
where $\beta>0$ controls the strength of reward tilting.
This reweighting increases the relative probability of higher-reward samples
while remaining within the support of the teacher distribution.
As $\beta\rightarrow0$, $p_0^\star$ reduces to
$p_{\mathrm T,0}$, recovering the standard teacher-matching objective.
Equivalently, ~\Eqref{eq:rydmd_target} arises as the solution to a
KL-regularized expected-reward objective; the derivation is provided in
Appendix~\ref{app:rydmd_global_derivation}.

\subsubsection{Local Reward-Tilted Teacher Target}

The global target in \Eqref{eq:rydmd_target} is not directly available during
DMD training. Instead, at each noisy student state, we sample multiple
teacher continuations and use their rewards to estimate how reward tilting
would shift the local teacher target.
During DMD training, the one-step generator predicts
$x_0^\theta=G_\theta(z,c)$, which is re-noised to
\begin{equation}
    x_\sigma
    =
    (1-\sigma)x_0^\theta+\sigma\epsilon,
    \qquad
    \epsilon\sim\mathcal N(0,I).
    \label{eq:rydmd_noisy_state}
\end{equation}
Starting from $x_\sigma$, a stochastic $K$-step teacher rollout produces
$x_0=\mathcal T_{\mathrm T}(x_\sigma,c;\eta)$,
where $\eta$ denotes the stochastic noise used during the rollout. These rollouts provide a Monte Carlo approximation of the conditional teacher
endpoint distribution
$q_{\mathrm T}^{K}(x_0\mid x_\sigma,c)$.
Extending the reward-tilting principle in ~\Eqref{eq:rydmd_target} to this
conditional distribution, we define a reward-reweighted rollout distribution as
\begin{equation}
    q_{\mathrm T,\beta}^{K}(x_0\mid x_\sigma,c)
    \propto
    q_{\mathrm T}^{K}(x_0\mid x_\sigma,c)
    \exp\!\bigl(\beta r(x_0,c)\bigr).
    \label{eq:rydmd_local_tilt}
\end{equation}
We estimate the resulting target shift using $J$ independent teacher
rollouts $\{x_0^{(j)}\}_{j=1}^{J}$ with rewards
$r_j=r(x_0^{(j)},c)$. Let $\pi_j$ denote the normalized reward weight of
the $j$-th rollout, with $\pi_j \propto \exp(\beta a_j)$ and
$a_j = (r_j-\bar r)/(s_r+\varepsilon)$,
where rewards are standardized within each rollout group for robustness to reward scale.
We then compute the original and reward-weighted teacher endpoint means as
\begin{equation}
    \widehat m_{\mathrm T}
    =
    J^{-1}\textstyle\sum_{j=1}^{J} x_0^{(j)},
    \qquad
    \widehat m_{\beta}
    =
    \textstyle\sum_{j=1}^{J} \pi_j x_0^{(j)}.
\end{equation}
and use their difference
\begin{equation}
    \delta m
    =
    \widehat m_{\beta}-\widehat m_{\mathrm T}
    \label{eq:rydmd_endpoint_shift}
\end{equation}
as the reward-induced shift of the local teacher target.
Crucially, $\delta m$ is estimated entirely from teacher samples and scalar
reward evaluations, without differentiating through the reward function. The normalized form of the local tilt, the Monte Carlo estimator, and
additional interpretation are provided in
Appendix~\ref{app:rydmd_local_derivation}.

\subsubsection{Reward-Yield DMD Update}

The reward-weighted rollouts tell us how the teacher endpoint target should move to favor higher-reward outputs. We incorporate this shift directly into the teacher term of DMD, while leaving the remaining distribution-matching procedure unchanged. Specifically, we convert the estimated endpoint displacement into a correction of the teacher term:
\begin{equation}
    s_{\mathrm T}^{\mathrm{RY}}
    =
    s_{\mathrm T}
    +
    \lambda_{\mathrm{RY}}
    \tfrac{\delta m}{\max(\sigma,\varepsilon_\sigma)},
    \label{eq:rydmd_update}
\end{equation}
where $\lambda_{\mathrm{RY}}$ controls the reward-correction
strength and $\varepsilon_\sigma$ stabilizes the update at
small noise levels.
The conversion from the endpoint displacement to the teacher
correction is provided in
Appendix~\ref{app:rydmd_update_derivation}.

\section{Experiments}
\label{sec:experiments}


\subsection{Comparison with State-of-the-Art VSR Methods}
\label{sec:exp_sota}

\begin{figure}[H]
    \centering
    \includegraphics[width=0.95\linewidth]{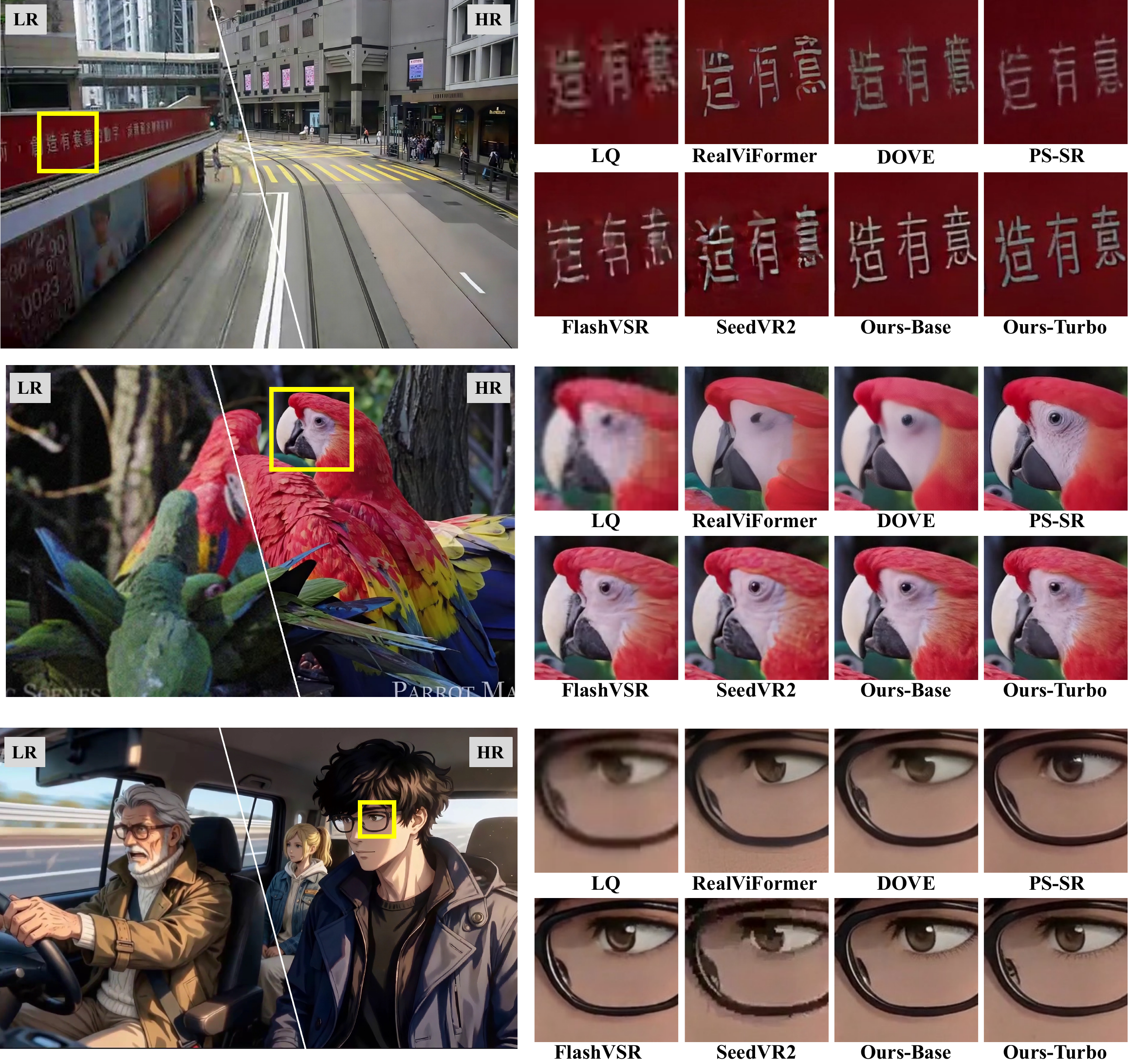}
    \caption{
    Qualitative comparison with representative VSR methods.
    Zoomed regions highlight the reconstruction of text, fine textures, and facial details.
    }
    \label{fig:vsr_visual_comparison}
\end{figure}

\begin{wraptable}{r}{0.52\textwidth}
\vspace{-1.5em}
\centering
\caption{\textbf{Human preference study.}
Preference margins relative to Ours-Base.}
\label{tab:user_study}
\vspace{-1.0em}
{\normalsize
\setlength{\tabcolsep}{4.0pt}
\renewcommand{\arraystretch}{1.05}
\begin{tabular}{lccc}
\toprule
Method & Fidelity & Quality & Overall \\
\midrule
RealViFormer
& $-70\%$ & $-72\%$ & $-73\%$ \\
DOVE
& $-68\%$ & $-68\%$ & $-70\%$ \\
DLoRAL
& $-38\%$ & $-28\%$ & $-35\%$ \\
SeedVR2
& $-53\%$ & $-55\%$ & $-56\%$ \\
PS-SR
& $-38\%$ & $-38\%$ & $-41\%$ \\
FlashVSR
& $-15\%$ & $-10\%$ & $-15\%$ \\
\midrule
Ours-Turbo
& $-7\%$ & $-5\%$ & $-6\%$ \\
\textbf{Ours-Base}
& \textbf{$0\%$} & \textbf{$0\%$} & \textbf{$0\%$} \\
\bottomrule
\end{tabular}
}
\vspace{-10pt}
\end{wraptable}
We evaluate on synthetic benchmarks, including YouHQ40~\citep{zhou2024upscale}, SPMCS~\citep{yi2019detail}, and RealisVideo-4K~\citep{zhao2025realisvsr}, as well as the real-world VideoLQ~\citep{chan2022realbasicvsr} and AI-generated AIGC35 benchmarks. Table~\ref{tab:vsr_sota} compares FlashVSR+ with representative VSR methods, including RealViFormer~\citep{zhang2024realviformer}, DOVE~\citep{chen2025dove}, SeedVR2-3B~\citep{wang2026seedvr2}, PS-SR~\citep{wu2026ps}, and FlashVSR~\citep{zhuang2026flashvsr}.
We report PSNR, SSIM, and LPIPS~\citep{zhang2018lpips} on datasets with ground truth, and MUSIQ~\citep{ke2021musiq}, CLIPIQA~\citep{wang2023clipiqa}, NIQE~\citep{mittal2013niqe}, and DOVER~\citep{wu2023dover} on all datasets. FlashVSR+ achieves strong overall restoration quality, with qualitative results in Fig.~\ref{fig:vsr_visual_comparison} showing better preservation of text, fine textures, and natural details with fewer spurious artifacts.
We further conduct a blinded pairwise user study following the GSB protocol used in ~\citep{wang2026seedvr2}. Ours-Base serves as the anchor, and each method pair is evaluated on 20 videos randomly sampled from RealisVideo-4K, VideoLQ, and AIGC35 benchmarks. Participants compare the two outputs in terms of visual fidelity, perceptual quality, and overall preference. Table~\ref{tab:user_study} reports the resulting preference margins relative to Ours-Base. Ours-Base is preferred over all compared methods, while Ours-Turbo remains close despite its higher inference efficiency.

\definecolor{firstbg}{RGB}{255,205,205}   
\definecolor{secondbg}{RGB}{255,225,180}  
\definecolor{thirdbg}{RGB}{255,245,180}   

\newcommand{\first}[1]{\cellcolor{firstbg}\textbf{#1}}
\newcommand{\second}[1]{\cellcolor{secondbg}\underline{#1}}
\newcommand{\third}[1]{\cellcolor{thirdbg}\textit{#1}}

\begin{table}[t]
\caption{
Quality comparison with state-of-the-art VSR methods under varying inference costs.
Top-1/2/3 results are highlighted in red/orange/yellow.
}

\label{tab:vsr_sota}
\centering
\flashtableformat
\setlength{\tabcolsep}{1.8pt}
\begin{tabular}{llccccccc}
\toprule
Dataset & Metric
& \shortstack{RealViFormer} & DOVE
& \shortstack{SeedVR2-3B} & PS-SR & FlashVSR
& \shortstack{Ours-Base} & \shortstack{Ours-Turbo} \\
\midrule

\multirow{7}{*}{YouHQ40}
& PSNR $\uparrow$
& \third{23.50} & 20.91 & 23.34 & 22.35 & 22.53
& \first{23.64} & \second{23.51} \\

& SSIM $\uparrow$
& 0.6051 & 0.5532 & \second{0.6206} & 0.5899 & 0.5787
& \first{0.6250} & \third{0.6097} \\

& LPIPS $\downarrow$
& 0.3528 & 0.3514 & 0.2768 & 0.3122 & \second{0.2721}
& \first{0.2698} & \third{0.2728} \\

& MUSIQ $\uparrow$
& 60.36 & 60.82 & 57.17 & \first{66.20} & \second{65.91}
& \third{64.71} & 63.91 \\

& CLIPIQA $\uparrow$
& 0.3823 & 0.3964 & 0.3703 & \first{0.5063} & \second{0.4701}
& 0.4257 & \third{0.4322} \\

& NIQE $\downarrow$
& \second{3.490} & 4.581 & 4.213 & 3.72 & \first{3.385}
& \third{3.560} & 3.737 \\

& DOVER $\uparrow$
& 10.42 & 10.97 & 12.23 & 12.93 & \first{13.47}
& \third{13.33} & \second{13.37} \\

\midrule

\multirow{7}{*}{SPMCS}
& PSNR $\uparrow$
& \second{22.64} & \first{23.09} & 22.24 & 21.16 & 21.35
& 22.34 & \third{22.43} \\

& SSIM $\uparrow$
& 0.5651 & \first{0.6109} & \second{0.5867} & 0.5452 & 0.5398
& \third{0.5714} & 0.5654 \\

& LPIPS $\downarrow$
& 0.3368 & 0.2738 & \first{0.2521} & 0.3240 & \third{0.2680}
& 0.2700 & \second{0.2647} \\

& MUSIQ $\uparrow$
& 61.91 & 68.93 & 65.63 & 69.47 & \third{69.59}
& \second{70.09} & \first{70.13} \\

& CLIPIQA $\uparrow$
& 0.3416 & 0.4605 & \third{0.4633} & \second{0.4650} & 0.4618
& \first{0.4753} & 0.4546 \\

& NIQE $\downarrow$
& 3.781 & 4.234 & 3.601 & 3.971 & \first{3.389}
& \second{3.401} & \third{3.411} \\

& DOVER $\uparrow$
& 7.428 & 8.352 & 8.462 & 8.907 & \third{9.250}
& \second{9.676} & \first{9.811} \\

\midrule

\multirow{4}{*}{Realis-4K}
& PSNR $\uparrow$
& 30.81 & 23.26 & \third{30.88} & 28.35 & 30.84
& \second{31.23} & \first{32.06} \\

& SSIM $\uparrow$
& 0.8240 & 0.7841 & 0.8206 & 0.8517 & \third{0.8596}
& \second{0.8639} & \first{0.8724} \\

& LPIPS $\downarrow$
& \second{0.1631} & 0.213 & 0.1929 & 0.1970 & 0.1671
& \third{0.1639} & \first{0.1625} \\

& DOVER $\uparrow$
& 11.35 & 12.60 & 12.93 & 12.20 & \third{13.27}
& \first{13.43} & \second{13.30} \\

\midrule
\midrule

\multirow{4}{*}{VideoLQ}
& NIQE $\downarrow$
& \second{4.485} & 4.915 & 5.114 & 5.00 & \first{4.309}
& \third{4.619} & 4.784 \\

& MUSIQ $\uparrow$
& \second{49.40} & 45.79 & 38.70 & 45.34 & \first{49.70}
& \third{47.64} & 46.93 \\

& CLIPIQA $\uparrow$
& \second{0.3117} & 0.3009 & 0.233 & \third{0.3066} & \first{0.3140}
& 0.2939 & 0.3013 \\

& DOVER $\uparrow$
& 7.217 & 7.542 & 5.955 & 7.769 & \third{7.811}
& \first{8.045} & \second{7.827} \\

\midrule

\multirow{4}{*}{AIGC35}
& NIQE $\downarrow$
& 4.983 & 5.173 & \first{4.816} & 4.923 & \third{4.888}
& \second{4.851} & 5.014 \\

& MUSIQ $\uparrow$
& 48.08 & 48.78 & 47.40 & \third{50.08} & \first{50.83}
& \second{50.30} & 49.53 \\

& CLIPIQA $\uparrow$
& 0.3871 & \second{0.4130} & 0.3945 & \first{0.4546} & \third{0.4058}
& 0.4012 & 0.4001 \\

& DOVER $\uparrow$
& 10.91 & 12.21 & 11.86 & 11.76 & \second{12.63}
& \first{13.01} & \third{12.57} \\

\bottomrule
\end{tabular}
\end{table}

\begin{figure}[h]
\caption{Efficiency comparison on 89-frame videos. (a) GPU time and peak memory on NVIDIA H200 equivalent GPU. (b) End-to-end wall-clock inference time on NVIDIA RTX 4090, including CPU KV-cache offloading and asynchronous prefetching overhead.}
\label{fig:vsr_efficiency}
\centering

\begin{minipage}[t]{0.715\linewidth}
\vspace{0pt}
\centering
{\small (a) NVIDIA H200 equivalent GPU}\par\vspace{0.2em}
{\small
\setlength{\tabcolsep}{3.5pt}
\begin{tabular}{l c ccc ccc}
\toprule
\multirow{2}{*}{Method} & \multirow{2}{*}{\shortstack{Backbone\\Params (B)}} & \multicolumn{3}{c}{GPU Time (s) $\downarrow$} & \multicolumn{3}{c}{Peak Memory (GB) $\downarrow$} \\
\cmidrule(lr){3-5} \cmidrule(lr){6-8}
& & 1080p & 2K & 4K & 1080p & 2K & 4K \\
\midrule
DOVE          & 5.0 & 89.38 & 165.3 & 374.7 & 26.54 & 26.22 & 26.56 \\
PS-SR         & 1.3 & 228.4 & 1123.1 & 1751.4 & 47.79 & 50.63 & 50.63 \\
SeedVR2       & 3.0 & 51.46 & 157.8$^{\dagger}$ & 410.5$^{\dagger}$ & 111.21 & 49.57$^{\dagger}$ & 51.76$^{\dagger}$ \\
\midrule
Ours-Base$^{\ddagger}$ & 1.3 & 7.91 & 17.28 & 46.34 & 13.11 & 21.29 & 45.00 \\
Ours-Turbo & 1.3 & \textbf{6.63} & \textbf{12.31} & \textbf{30.59} & \textbf{8.72} & \textbf{13.23} & \textbf{26.33} \\
\bottomrule
\end{tabular}
}
\end{minipage}\hfill
\begin{minipage}[t]{0.265\linewidth}
\vspace{0pt}
\centering
{\small (b) NVIDIA RTX 4090}\par\vspace{0.2em}
\includegraphics[width=\linewidth]{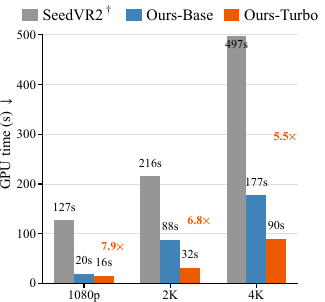}
\end{minipage}

\vspace{0.4em}
\begin{minipage}{\textwidth}
\footnotesize
$^{\dagger}$ SeedVR2 full-frame inference OOMs so we use tiled inference.
$^{\ddagger}$ Ours-Base shares the same architecture and inference procedure as FlashVSR on H200 equivalent GPU. FlashVSR's attention operators are unsupported on RTX 4090.
\end{minipage}

\end{figure}

\begin{wrapfigure}{r}{0.42\columnwidth}
\vspace{-0.5em}
\centering
\includegraphics[width=\linewidth]{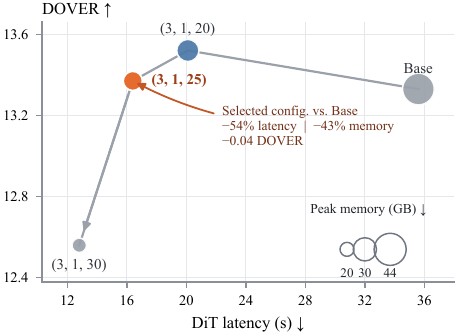}
\vspace{-1.4em}
\caption{
\textbf{Quality and efficiency of trained hybrid attention.}
The $(3,1,25)$ configuration achieves the best balance between DiT cost and restoration quality.
}
\label{fig:hybrid_ablation}
\vspace{-3.6em}
\end{wrapfigure}
We achieve substantial efficiency gains for high-resolution VSR. We evaluate computational efficiency in Figure~\ref{fig:vsr_efficiency} using 89-frame videos at 1080p, 2K, and 4K resolutions on NVIDIA H200 equivalent GPU and consumer RTX 4090 GPUs. We substantially reduce both inference latency and peak memory compared with the base model, enabling high-resolution restoration with much lower computational overhead. In particular, with additional memory optimizations, we enable 4K VSR inference on a single RTX 4090; implementation details are provided in Appendix~\ref{app:implement_4090}.

\subsection{Efficient VSR with Hybrid Attention}
\label{sec:exp_hybrid}

We evaluate hybrid attention both before and after adaptation. The training-free setting tests how much of the pretrained attention can be localized directly, while the adapted setting measures the resulting quality and efficiency after training the modified architecture.

We first consider the training-free setting. We use $(3,1,15)$ as a conservative reference and vary one factor at a time. Since the attention pattern is modified without retraining, we only count clear structural failures as artifacts, including block or grid-shaped corruption and severe temporal ghosting; minor changes in sharpness or texture are not considered failures. The pretrained model tolerates $(3,1,15)$, but reducing the spatial or temporal context, or increasing the number of localized blocks, quickly introduces artifacts. In particular, $(1,1,15)$, $(3,0,15)$, and configurations with $b\geq20$ fail under direct replacement, motivating explicit adaptation of the modified attention structure.

We next adapt these hybrid-attention configurations. As shown in Fig.~\ref{fig:hybrid_ablation}, adaptation substantially extends the usable range of hybrid attention. Increasing the number of local-attention blocks reduces latency and peak memory while preserving restoration quality up to $(3,1,25)$, whereas further localization to $(3,1,30)$ causes a clear quality drop. We therefore adopt $(3,1,25)$ as the default configuration. The more restrictive $(1,1,15)$ and $(3,0,15)$ configurations still exhibit visible artifacts after adaptation and are therefore excluded from the figure.

\subsection{Reward-Guided VSR Post-Training}
\label{sec:exp_rydmd}

\begin{wrapfigure}{r}{0.6\textwidth}
\vspace{-1.0em}
\centering
\includegraphics[width=\linewidth]{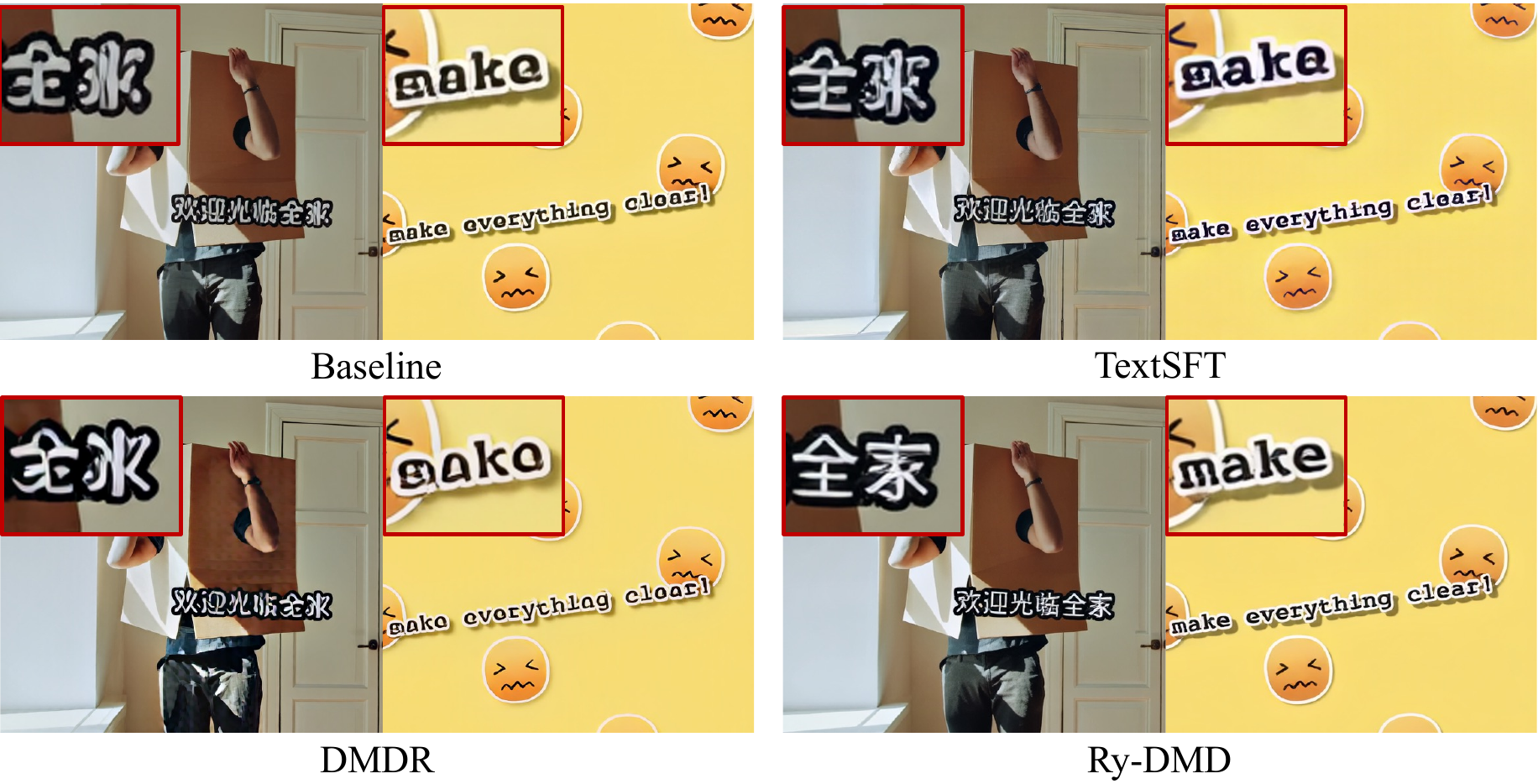}
\vspace{-2em}
\caption{
\textbf{Qualitative comparison of OCR-guided VSR post-training.}
Ry-DMD restores more recognizable text, while DMDR often introduces artifacts.
}
\label{fig:ocr_reward_visual}
\vspace{-2em}
\end{wrapfigure}

We evaluate Ry-DMD on OCR-guided VSR post-training, a task requiring fine-grained visual fidelity~\citep{yuan2026detached},
using a synthetic text-focused dataset with annotated text ground truth; details of the data generation pipeline
are provided in Appendix~\ref{app:ocr_data}. All methods start from the same
pretrained one-step VSR model.

Following the standard practice in Flow-GRPO~\citep{liu2025flowgrpo}, given target text $y$ and the text $\hat y$ recognized by
PaddleOCR~\citep{cui2025paddleocr30technicalreport}, we compute the OCR reward as
$
r_{\mathrm{OCR}}
=
1-\frac{\min(d_{\mathrm{Lev}}(\hat y,y),|y|)}{|y|},
$
where both strings are stripped of spaces before evaluation.

\begin{wraptable}{r}{0.43\columnwidth}
\centering
\vspace{-1.5em}
\caption{
\textbf{Quantitative results of OCR-guided post-training.}
}
\label{tab:ocr_reward_comparison}
\vspace{-0.8em}
\small
\setlength{\tabcolsep}{1.3pt}
\renewcommand{\arraystretch}{1.05}
\resizebox{\linewidth}{!}{%
\begin{tabular}{@{}lccccc@{}}
\toprule
Method & OCR $\uparrow$ & PSNR $\uparrow$ & LPIPS $\downarrow$
& MUSIQ $\uparrow$ & DOVER $\uparrow$ \\
\midrule
Baseline & 0.624 & 23.94 & 0.246 & 75.35 & 9.79 \\
TextSFT  & 0.656 & 22.02 & 0.295 & \textbf{76.41} & 10.58 \\
DMDR     & 0.527 & 19.01 & 0.374 & 74.17 & 8.48 \\
Ry-DMD   & \textbf{0.776} & \textbf{24.86} & \textbf{0.240}
& 75.80 & \textbf{10.90} \\
\bottomrule
\end{tabular}}
\end{wraptable}

We compare four post-training strategies. {Baseline} denotes the original pretrained model without further adaptation. {TextSFT} continues standard DMD training on the text-focused dataset, testing whether task-specific data alone is sufficient. {DMDR}~\citep{jiang2025dmdr} directly introduces an OCR optimization objective into DMD training; since the OCR reward is non-differentiable, we use a CTC-based differentiable surrogate and backpropagate through the OCR model. Instead, Ry-DMD directly uses the scalar OCR reward without requiring its gradient. As shown in Table~\ref{tab:ocr_reward_comparison}, Ry-DMD achieves the strongest OCR improvement while preserving restoration quality. The qualitative results in Fig.~\ref{fig:ocr_reward_visual} further show clearer text restoration with fewer artifacts.

\subsection{Generalization to Text-to-Image Generation}
\label{sec:exp_t2i}

We further evaluate whether Ry-DMD generalizes beyond restoration by applying it to one-step SDXL text-to-image generation. We initialize the student from a one-step DMD2 model~\citep{yin2024dmd2} distilled from SDXL~\citep{podell2024sdxl}. Training uses OCR prompts from Flow-GRPO~\citep{liu2025flowgrpo} together with a disjoint training split of DrawBench~\citep{saharia2022imagen}. We optimize a combined reward consisting of OCR and PickScore. Ry-DMD relies only on scalar reward evaluations and does not backpropagate through either scorer.

\begin{table}[!ht]
\vspace{-0.1em}
\caption{
Reward-guided text-to-image generation on SDXL.
}
\label{tab:t2i_reward_comparison}
\vspace{-0.8em}
\centering
\flashtableformat
\begin{tabular}{lcccccc}
\toprule
Model & NFE $\downarrow$ & OCR $\uparrow$ & PickScore $\uparrow$ & HPSv2 $\uparrow$ & ImageReward $\uparrow$ & Aesthetic $\uparrow$ \\
\midrule
SDXL                         & 50 & 0.125 & 22.2 & 27.5 & 0.87 & \textbf{5.48} \\
DMD2                         & 4  & 0.098 & 22.5 & 29.2 & 0.83 & 5.47 \\
One-step DMD2                & 1  & 0.087 & 22.0 & 26.8 & 0.67 & 5.39 \\
Ry-DMD (ours)                & 1  & \textbf{0.241} & \textbf{22.7} & \textbf{29.6} & \textbf{1.02} & {5.44} \\
\bottomrule
\end{tabular}
\vspace{-0.6em}
\end{table}

\begin{figure}[h]
    \centering
    \includegraphics[width=0.8\linewidth]{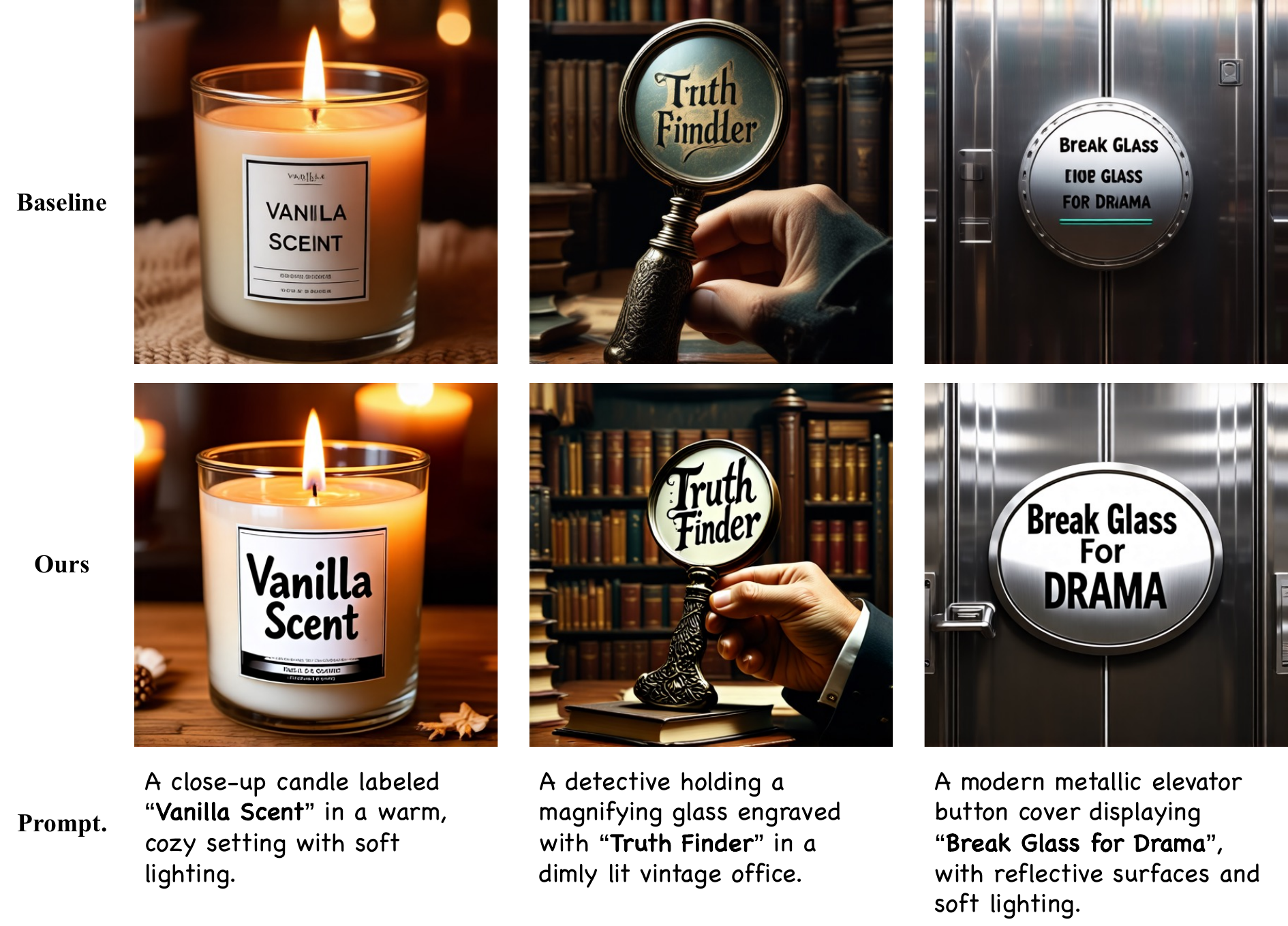}
    \caption{
    \textbf{Qualitative comparison of Ry-DMD on 4-step SD3.5-M.}
    Ry-DMD substantially improves text correctness and legibility while
    preserving the intended scene content. Beyond text rendering, the
    resulting images often exhibit cleaner structures, more coherent
    materials and lighting, and improved overall visual quality.
    }
    \label{fig:rydmd_fourstep_visual}
\end{figure}

We report PickScore~\citep{kirstain2023pickapic}, HPSv2~\citep{wu2023hpsv2}, ImageReward~\citep{xu2023imagereward}, and the LAION aesthetic predictor~\citep{schuhmann2022laion5b}. As shown in Table~\ref{tab:t2i_reward_comparison}, Ry-DMD improves the OCR score from 0.087 to 0.241, surpassing the 50-step teacher, while maintaining or improving complementary quality metrics. Ry-DMD also generalizes to four-step SD3.5-M, improving OCR accuracy from 0.501 to 0.824 without increasing NFE. As shown in Fig.~\ref{fig:rydmd_fourstep_visual}, Ry-DMD improves text rendering while preserving scene content and overall visual quality. Additional results and hyperparameter analysis are provided in Appendices~\ref{app:rydmd_hyperparams} and~\ref{app:rydmd_fourstep}.

\subsection{Rethinking DMD Scaling for One-Step VSR}
\label{sec:dmd_scaling}

\begin{wrapfigure}{r}{0.42\columnwidth}
    \vspace{-2.0em}
    \centering
    \includegraphics[width=\linewidth]{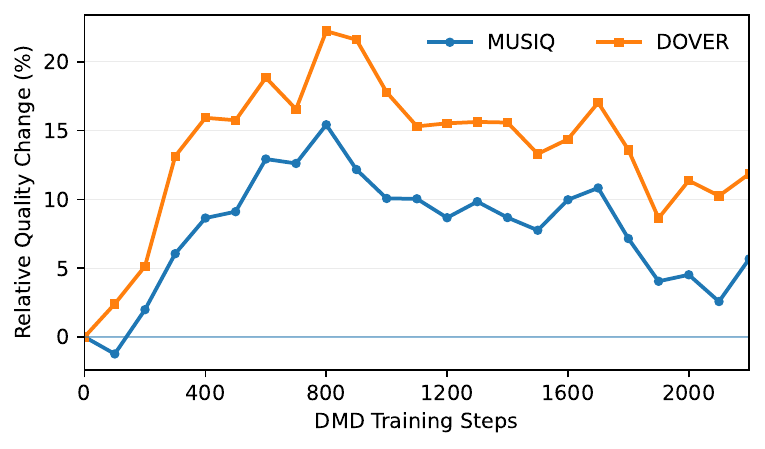}
    \vspace{-2.5em}
    \caption{
    \textbf{Ablation on DMD steps.}
    }
    \label{fig:dmd_scaling_curve}
    \vspace{-1.5em}
\end{wrapfigure}

To understand how to effectively scale up one-step VSR training, we first investigate whether allocating a larger optimization budget can consistently improve DMD performance. To reduce domain bias, we construct a fixed held-out evaluation
set by sampling from multiple benchmarks, including AIGC, YouHQ, and VideoLQ,
covering both synthetic and real-world content with diverse degradations.
Intermediate checkpoints are evaluated using MUSIQ and DOVER.

As shown in Fig.~\ref{fig:dmd_scaling_curve}, both metrics improve rapidly
during early DMD training. However, the gains do not persist: after an early
peak, performance becomes non-monotonic and can decline with continued
optimization. Simply increasing the number of DMD updates therefore does not
reliably yield a better one-step VSR model.

\begin{wraptable}{r}{0.4\columnwidth}
    \vspace{-1.5em}
    \centering
    \caption{\textbf{Ablation on student initialization and teacher capacity.}}
    \label{tab:dmd_factors}
    \vspace{-0.5em}
    {\small
    \setlength{\tabcolsep}{1.8pt}
    \renewcommand{\arraystretch}{0.92}
    \begin{tabular}{@{}cccc@{}}
        \toprule
        Init. & Teacher & MUSIQ $\uparrow$ & DOVER $\uparrow$ \\
        \midrule
        Base   & 1.3B & 45.87 & 8.512 \\
        Strong & 1.3B & 48.20 & 8.738 \\
        Base   & 14B  & 51.76 & 9.144 \\
        Strong & 14B  & \textbf{52.78} & \textbf{9.417} \\
        \bottomrule
    \end{tabular}
    }
    \vspace{-1.8em}
\end{wraptable}

We next examine two other components of the distillation setup: student
initialization and teacher capacity. As shown in Table~\ref{tab:dmd_factors},
both a stronger initialization and a larger teacher consistently improve the
final student, with their combination performing best. These results indicate
that effective DMD scaling depends on the overall distillation setup rather
than optimization duration alone.

\section{Conclusion}
\label{sec:conclusion}

We introduce FlashVSR+, combining hybrid spatiotemporal attention with Ry-DMD for efficient and reward-guided one-step VSR.
The framework improves the quality efficiency trade-off and enables spatially untiled 4K inference on a single 24GB RTX 4090.
Ry-DMD enables black-box reward optimization without reward backpropagation, with benefits demonstrated in both VSR and text-to-image generation.

\bibliography{references}
\bibliographystyle{plainnat}

\newpage
\appendix

\section{Training Details}
\label{app:train_details} 
In Stage 1, we train the 14B full-attention teacher for 4 days using LoRA with rank 384, AdamW with $(\beta_1,\beta_2)=(0.9,0.999)$, a learning rate of $2\times10^{-5}$, and a batch size of 16. In Stage 2, we initialize the hybrid-attention student from pretrained FlashVSR stage-2 checkpoint and perform full-parameter training for 2 days using AdamW with $(\beta_1,\beta_2)=(0.9,0.999)$, a learning rate of $1\times10^{-5}$, a batch size of 32, and EMA. In Stage 3, we perform full-parameter one-step distillation for 600 steps using the 14B teacher, AdamW with $(\beta_1,\beta_2)=(0,0.999)$, a learning rate of $1\times10^{-5}$, a batch size of 16, and EMA. Training uses a filtered coreset together with the text-SFT augmentation. For Ry-DMD, we set $\beta=0.1$, $\lambda_{\mathrm{RY}}=0.2$, $J=4$, and $K=3$.

\section{RTX 4090 Implementation}
\label{app:implement_4090}

The attention operators used in our H200 equivalent GPU implementation are not directly supported on RTX 4090 GPUs. We therefore implement a lightweight Triton adaptation based on FlashAttention-2 to support the same hybrid-attention computation on Ada GPUs. 

For 4K inference, the main memory bottleneck comes from the persistent KV cache of the retained global-attention layers. We selectively offload these KV caches to pinned CPU memory and asynchronously transfer them on a dedicated CUDA stream. Before the corresponding layer is executed in the next streaming chunk, the cache is prefetched back to GPU. This substantially reduces persistent GPU memory usage while overlapping most data transfers with computation. Importantly, this adaptation does not introduce spatial tiling: each frame is processed at its full spatial resolution, while only the KV-cache residency is changed. This enables spatially untiled 4K inference on a single 24GB RTX 4090.

\section{Ry-DMD Hyperparameter Analysis}
\label{app:rydmd_hyperparams}

We study the sensitivity of Ry-DMD to four key hyperparameters on SDXL:
the reward temperature $\beta$, the correction weight
$\lambda_{\mathrm{RY}}$, the number of teacher rollouts $J$, and the rollout
depth $K$. Unless otherwise specified, we use
$\beta=0.1$, $\lambda_{\mathrm{RY}}=0.2$, $J=4$, and $K=3$.
We vary one parameter at a time while keeping the others fixed.
As shown in Table~\ref{tab:rydmd_hyperparams}, Ry-DMD is stable under moderate variations of its hyperparameters, while overly strong reward weighting or correction degrades performance.
\begin{table}[h]
\centering
\caption{
\textbf{Hyperparameter analysis of Ry-DMD on SDXL.}
We vary one hyperparameter at a time from the default configuration
$(\beta,\lambda_{\mathrm{RY}},J,K)=(0.1,0.2,4,3)$.
}
\label{tab:rydmd_hyperparams}
\setlength{\tabcolsep}{4.5pt}
\begin{tabular}{lcccccc}
\toprule
Setting
& $\beta$
& $\lambda_{\mathrm{RY}}$
& $J$
& $K$
& OCR $\uparrow$
& ImageReward $\uparrow$ \\
\midrule

Default
& 0.1 & 0.2 & 4 & 3
& 0.241 & 1.02 \\
\midrule

$\beta$
& 0.05 & -- & -- & --
& 0.212 & 0.98 \\
& 0.2 & -- & -- & --
& 0.236 & 1.01 \\
& 0.4 & -- & -- & --
& 0.203 & 0.92 \\
\midrule

$\lambda_{\mathrm{RY}}$
& -- & 0.05 & -- & --
& 0.201 & 0.98 \\
& -- & 0.1 & -- & --
& 0.237 & 1.01 \\
& -- & 0.4 & -- & --
& 0.198 & 0.89 \\
\midrule

$J$
& -- & -- & 2 & --
& 0.227 & 1.00 \\
& -- & -- & 8 & --
& 0.248 & 1.04 \\
\midrule

$K$
& -- & -- & -- & 2
& 0.239 & 1.01 \\
& -- & -- & -- & 4
& 0.239 & 1.03 \\
\bottomrule
\end{tabular}
\end{table}

\section{Extending Ry-DMD to Few-Step Generation}
\label{app:rydmd_fourstep}

We provide additional quantitative results for four-step SD3.5-M with OCR-guided post-training. As shown in Table~\ref{tab:rydmd_fourstep}, Ry-DMD improves OCR accuracy from 0.501 to 0.824 without increasing NFE.
Meanwhile, HPSv2, ImageReward, and Aesthetic scores all improve, indicating that better text rendering does not come at the expense of general image quality. Qualitative comparisons are provided in Fig.~\ref{fig:rydmd_fourstep_visual}.

\begin{table}[h]
    \centering
    \caption{
    Ry-DMD on 4-step SD3.5-M. Ry-DMD substantially improves OCR accuracy
    while preserving the original 4-step sampling budget and improving
    general image-quality and preference metrics.
    }
    \label{tab:rydmd_fourstep}
    \begin{tabular}{lccccc}
    \toprule
    Method & NFE $\downarrow$ & OCR $\uparrow$ & HPSv2 $\uparrow$
    & ImageReward $\uparrow$ & Aesthetic $\uparrow$ \\
    \midrule
    SD3.5-M (4-step) & 4 & 0.501 & 30.3 & 1.02 & 5.52 \\
    + Ry-DMD & 4 & \textbf{0.824} & \textbf{31.2}
    & \textbf{1.23} & \textbf{5.54} \\
    \bottomrule
    \end{tabular}
\end{table}

\section{Additional Derivations for Ry-DMD}
\label{app:rydmd_impl}

This section provides supporting derivations for Ry-DMD introduced in
Section~\ref{sec:rydmd}. We first show that the global reward-tilted target
arises from a KL-regularized expected-reward objective. We then derive the
local reward-tilted endpoint mean and its finite-rollout estimator, followed
by the connection between the resulting endpoint shift and the DMD teacher
correction.

\subsection{Derivation of the Reward-Tilted Target}
\label{app:rydmd_global_derivation}

For a fixed condition $c$, consider the KL-regularized expected-reward
objective
\begin{equation}
\mathcal J(q)
=
\mathbb E_{x_0\sim q(\cdot\mid c)}
\left[r(x_0,c)\right]
-
\frac{1}{\beta}
D_{\mathrm{KL}}
\left(
q(\cdot\mid c)
\,\|\, 
p_{\mathrm T,0}(\cdot\mid c)
\right),
\label{eq:app_rydmd_variational}
\end{equation}
subject to
$\int q(x_0\mid c)\,\mathrm d x_0=1$.
Introducing a Lagrange multiplier for the normalization constraint and
setting the functional derivative with respect to $q$ to zero gives
\begin{equation}
\log
\frac{
q^\star(x_0\mid c)
}{
p_{\mathrm T,0}(x_0\mid c)
}
=
\beta r(x_0,c)
+
\mathrm{const}.
\end{equation}
Therefore,
\begin{equation}
p_0^\star(x_0\mid c)
=
\frac{
p_{\mathrm T,0}(x_0\mid c)
\exp\!\left(\beta r(x_0,c)\right)
}{
Z(c)
},
\qquad
Z(c)
=
\mathbb E_{p_{\mathrm T,0}(\cdot\mid c)}
\left[
\exp\!\left(\beta r(x_0,c)\right)
\right],
\label{eq:app_rydmd_global_target}
\end{equation}
which recovers the reward-tilted target in
\Eqref{eq:rydmd_target}. The exponential tilt changes the relative density
of teacher-supported samples according to their rewards while preserving
the support of the teacher distribution. In the limit
$\beta\rightarrow0$,
$p_0^\star(\cdot\mid c)\rightarrow
p_{\mathrm T,0}(\cdot\mid c)$.

\subsection{Local Reward-Tilted Mean Estimation}
\label{app:rydmd_local_derivation}

During DMD training, reward tilting is applied locally to the conditional
endpoint distribution induced by stochastic teacher rollouts from the
current noisy state. For fixed $(x_\sigma,c)$, let
$q_{\mathrm T}^{K}(x_0\mid x_\sigma,c)$ denote the endpoint distribution
induced by a stochastic $K$-step teacher sampler. The normalized form of
\Eqref{eq:rydmd_local_tilt} is
\begin{equation}
q_{\mathrm T,\beta}^{K}(x_0\mid x_\sigma,c)
=
\frac{
q_{\mathrm T}^{K}(x_0\mid x_\sigma,c)
\exp\!\left(\beta r(x_0,c)\right)
}{
Z_\beta(x_\sigma,c)
},
\label{eq:app_local_tilt}
\end{equation}
where
\begin{equation}
Z_\beta(x_\sigma,c)
=
\mathbb E_{
x_0\sim q_{\mathrm T}^{K}(\cdot\mid x_\sigma,c)
}
\left[
\exp\!\left(\beta r(x_0,c)\right)
\right].
\end{equation}

The mean endpoint under the reward-tilted distribution is therefore
\begin{equation}
m_\beta(x_\sigma,c)
=
\frac{
\mathbb E_{
x_0\sim q_{\mathrm T}^{K}(\cdot\mid x_\sigma,c)
}
\left[
x_0\exp\!\left(\beta r(x_0,c)\right)
\right]
}{
\mathbb E_{
x_0\sim q_{\mathrm T}^{K}(\cdot\mid x_\sigma,c)
}
\left[
\exp\!\left(\beta r(x_0,c)\right)
\right]
},
\label{eq:app_tilted_mean}
\end{equation}
while the original conditional teacher mean is
\begin{equation}
m_{\mathrm T}(x_\sigma,c)
=
\mathbb E_{
x_0\sim q_{\mathrm T}^{K}(\cdot\mid x_\sigma,c)
}
[x_0].
\end{equation}
We denote the population-level displacement induced by reward tilting as
\begin{equation}
\Delta m
=
m_\beta(x_\sigma,c)
-
m_{\mathrm T}(x_\sigma,c).
\label{eq:app_exact_shift}
\end{equation}

In practice, we draw $J$ independent teacher rollouts
\begin{equation}
x_0^{(j)}
\overset{\mathrm{i.i.d.}}{\sim}
q_{\mathrm T}^{K}(\cdot\mid x_\sigma,c),
\qquad
j=1,\ldots,J,
\end{equation}
with rewards $r_j=r(x_0^{(j)},c)$.
Without reward standardization, the corresponding self-normalized Monte
Carlo estimates are
\begin{equation}
\widehat m_\beta
=
\sum_{j=1}^{J}\pi_jx_0^{(j)},
\qquad
\pi_j
=
\frac{
\exp(\beta r_j)
}{
\sum_{k=1}^{J}\exp(\beta r_k)
},
\label{eq:app_snis_mean}
\end{equation}
and
\begin{equation}
\widehat m_{\mathrm T}
=
\frac{1}{J}
\sum_{j=1}^{J}
x_0^{(j)}.
\end{equation}
The finite-rollout estimate of the reward-induced endpoint shift is thus
\begin{equation}
\delta m
=
\widehat m_\beta-\widehat m_{\mathrm T},
\label{eq:app_mc_shift}
\end{equation}
corresponding to \Eqref{eq:rydmd_endpoint_shift}. 
The expression above gives the unstandardized estimator of
$\Delta m$. In practice, we compute $\delta m$ using the
group-standardized weights below, as in
\Eqref{eq:rydmd_endpoint_shift}.

\paragraph{Reward standardization.}
In the practical algorithm, rewards are standardized within each rollout
group as
\begin{equation}
a_j
=
\frac{r_j-\bar r}{s_r+\varepsilon},
\qquad
\pi_j
=
\frac{
\exp(\beta a_j)
}{
\sum_{k=1}^{J}\exp(\beta a_k)
}.
\label{eq:app_standardized_weights}
\end{equation}
Centering by $\bar r$ does not affect the normalized exponential weights,
since the common multiplicative factor cancels. Dividing by
$s_r+\varepsilon$ is equivalent to using a rollout-dependent effective
temperature
\begin{equation}
\beta_{\mathrm{eff}}
=
\frac{\beta}{s_r+\varepsilon},
\end{equation}
which reduces sensitivity to the numerical scale of different reward
functions while preserving their within-group ordering. The standardized weights therefore implement a group-adaptive
tilt, while the fixed-$\beta$ population formulas above describe the unstandardized case.

\paragraph{Small-$\beta$ interpretation.}
For additional intuition, a first-order expansion of
\Eqref{eq:app_tilted_mean} gives
\begin{equation}
\Delta m
=
\beta
\operatorname{Cov}_{
q_{\mathrm T}^{K}(\cdot\mid x_\sigma,c)
}
\left(
x_0,
r(x_0,c)
\right)
+
\mathcal O(\beta^2).
\label{eq:app_covariance_interpretation}
\end{equation}
Thus, in the small-$\beta$ regime, reward tilting moves the conditional
teacher endpoint mean along variations in $x_0$ that covary positively with
the reward.

\subsection{From Endpoint Shift to the DMD Teacher Correction}
\label{app:rydmd_update_derivation}

We finally derive how the endpoint displacement estimated above translates into the reward correction in \Eqref{eq:rydmd_update}. In this section, $s_{\mathrm T}$ denotes the parameterized teacher term used in our DMD update. For $\sigma>0$, it is related to its associated clean-endpoint target$\mu_{\mathrm T}(x_\sigma,c)$ by
\begin{equation}
s_{\mathrm T}(x_\sigma,c)
=
\frac{
\mu_{\mathrm T}(x_\sigma,c)-x_\sigma
}{
\sigma
}.
\label{eq:app_dmd_teacher_score}
\end{equation}
Ry-DMD modifies the clean teacher target using the estimated reward-induced
endpoint displacement:
\begin{equation}
\mu_{\mathrm T}^{\mathrm{RY}}(x_\sigma,c)
=
\mu_{\mathrm T}(x_\sigma,c)
+
\lambda_{\mathrm{RY}}\delta m,
\label{eq:app_corrected_endpoint}
\end{equation}
where $\lambda_{\mathrm{RY}}$ controls the correction strength.
Substituting this target into \Eqref{eq:app_dmd_teacher_score} yields
\begin{equation}
\begin{aligned}
s_{\mathrm T}^{\mathrm{RY}}(x_\sigma,c)
&=
\frac{
\mu_{\mathrm T}(x_\sigma,c)
+
\lambda_{\mathrm{RY}}\delta m
-
x_\sigma
}{
\sigma
}
\\
&=
s_{\mathrm T}(x_\sigma,c)
+
\lambda_{\mathrm{RY}}
\frac{\delta m}{\sigma}.
\end{aligned}
\label{eq:app_corrected_target}
\end{equation}
Thus, under the above parameterization, the factor $1/\sigma$
follows directly from converting a clean-endpoint displacement into a displacement of the teacher term.
For numerical stability at small noise levels, we use
\begin{equation}
d_{\mathrm{RY}}
=
\frac{
\delta m
}{
\max(\sigma,\varepsilon_\sigma)
},
\end{equation}
which gives the practical update
\begin{equation}
s_{\mathrm T}^{\mathrm{RY}}
=
s_{\mathrm T}
+
\lambda_{\mathrm{RY}}d_{\mathrm{RY}},
\end{equation}
equivalent to \Eqref{eq:rydmd_update}. Thus, Ry-DMD leaves the remaining DMD
optimization procedure unchanged and requires only sampled teacher endpoints
and scalar reward evaluations, without differentiating through the reward
function. Here, $\mu_{\mathrm T}$ remains the original teacher target,
and $\delta m$ is used as a rollout-estimated residual
correction; the construction does not require
$\mu_{\mathrm T}=m_{\mathrm T}$.

\section{Synthetic Text-OCR VSR Dataset}
\label{app:ocr_data}

To study OCR-guided post-training, we construct a synthetic text-focused VSR dataset from the same video corpus used in our VSR pipeline. Synthetic text is rendered onto decoded high-quality frames to form the target video, and the corresponding low-quality input is generated with the standard VSR degradation pipeline. Since the text is procedurally generated, its content and spatial annotations are known exactly.

\paragraph{Text rendering.}
Each sample contains one or two text instances sampled from a multilingual corpus covering English, Chinese, numeric strings, and several additional languages. We randomize font, size, color, outline, opacity, and shadow. Text remains temporally coherent across the video and follows simple motion patterns such as static placement, translation, scale/rotation changes, or partial motion outside the frame. Per-frame locations and visibility are retained as annotations.

\paragraph{Filtering and OCR supervision.}
We discard invalid renderings with unsupported characters, insufficient size, or negligible visible area. For OCR reward computation, only sufficiently visible annotated text regions are evaluated, and a small number of valid frames are sampled from each video for efficiency. All compared post-training methods use the same source videos, rendered targets, degradation process, and filtering rules.

Each sample additionally stores lightweight text metadata used for OCR supervision.
A schematic example is shown below:

\begin{lstlisting}
{
  "source_path": "/videos/example.mp4",
  "gt_video": "[F, 3, H, W]",
  "lq_video": "[F, 3, H, W]",
  "text_metadata": {
    "instances": [{
      "text": "EXAMPLE 42",
      "language": "en",
      "motion_type": "translate",
      "frames": [
        {"frame_index": 0,
         "bbox": [x0, y0, x1, y1],
         "visible_ratio": 1.0}
      ]
    }]
  }
}
\end{lstlisting}

\end{document}

%% file: math_commands.tex
\usepackage{amsmath,amsfonts,bm}

\def\eqref#1{eq.~\ref{#1}}
\def\Eqref#1{Eq.~\ref{#1}}

\def\1{\bm{1}}

\DeclareMathAlphabet{\mathsfit}{\encodingdefault}{\sfdefault}{m}{sl}
\SetMathAlphabet{\mathsfit}{bold}{\encodingdefault}{\sfdefault}{bx}{n}

